\documentclass{article}

\input{config.sty}
\input{glossary.sty}

\usepackage[accepted]{icml2025}

\icmltitlerunning{Targeted control of fast prototyping through domain-specific interface}

\begin{document}

\twocolumn[
    \icmltitle{Targeted control of fast prototyping through domain-specific interface}
    \icmlsetsymbol{equal}{*}
    
    \begin{icmlauthorlist}
    \icmlauthor{Yu-Zhe Shi}{equal,pku,cse}
    \icmlauthor{Mingchen Liu}{equal,hust}
    \icmlauthor{Hanlu Ma}{emia}
    \icmlauthor{Qiao Xu}{pku}
    \icmlauthor{Huamin Qu}{cse,emia}
    \icmlauthor{Kun He}{hust}\\
    \icmlauthor{Lecheng Ruan}{pku}
    \icmlauthor{Qining Wang}{pku}
    \end{icmlauthorlist}
    
    \icmlaffiliation{pku}{Department of Advanced Manufacturing and Robotics, College of Engineering, Peking University, Beijing, China}
    \icmlaffiliation{cse}{Department of Computer Science and Engineering, The Hong Kong University of Science and Technology, Hong Kong SAR}
    \icmlaffiliation{hust}{School of Computer Science and Technology, Huazhong University of Science and Technology, Wuhan, China}
    \icmlaffiliation{emia}{Division of Emerging Interdisciplinary Areas, The Hong Kong University of Science and Technology, Hong Kong SAR}
    
    \icmlcorrespondingauthor{Lecheng Ruan}{ruanlecheng@ucla.edu}
    \icmlcorrespondingauthor{Qining Wang}{qiningwang@pku.edu.cn}
    
    \icmlkeywords{Bayesian Generalization, Rational Analysis, Natural Image Statistics}
    \vskip 0.3in
]

\printAffiliationsAndNotice{\icmlEqualContribution}

\begin{abstract}
Industrial designers have long sought a natural and intuitive way to achieve the targeted control of prototype models---using simple natural language instructions to configure and adjust the models seamlessly according to their intentions, without relying on complex modeling commands. While Large Language Models have shown promise in this area, their potential for controlling prototype models through language remains partially underutilized. This limitation stems from gaps between designers' languages and modeling languages, including mismatch in abstraction levels, fluctuation in semantic precision, and divergence in lexical scopes. To bridge these gaps, we propose an interface architecture that serves as a medium between the two languages. Grounded in design principles derived from a systematic investigation of fast prototyping practices, we devise the interface's operational mechanism and develop an algorithm for its automated domain specification. Both machine-based evaluations and human studies on fast prototyping across various product design domains demonstrate the interface's potential to function as an auxiliary module for Large Language Models, enabling precise and effective targeted control of prototype models.
\end{abstract}

\begin{figure}[t!]
    \centering
    \includegraphics[width=\linewidth]{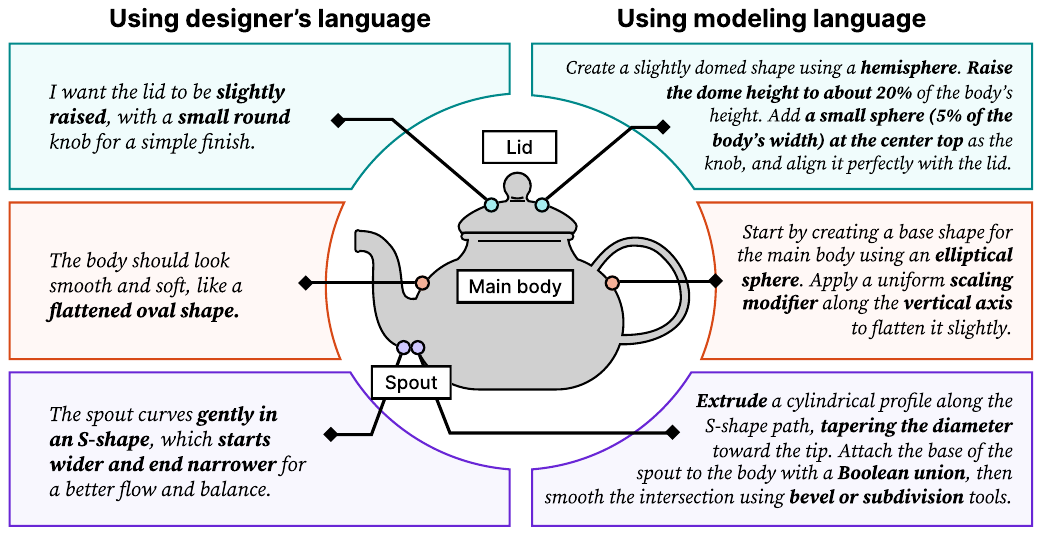}
    \vspace{-\baselineskip}
    \caption{\textbf{Designers' language \vs modeling language for fast prototyping.} This comparison clearly demonstrates that the designers' language for teapot design is significantly more intuitive than its modeling counterpart, providing a focused creation flow.}
    \label{fig:motivation}
\end{figure}

\section{Introduction}

Creating physical prototypes of the target products in a rapid iteration fashion, known as \emph{fast prototyping}~\citep{burns1993automated,hallgrimsson2012prototyping}, is one of the most critical processes in professional industrial design~\citep{camburn2017design}. It helps maintain project timelines, alleviates design fixation, and enables product testing against clear criteria. Unlike production-ready modeling conducted by engineers~\citep{barnhill1992geometry}, which focuses on refining prototypes into master models for mass-production, fast prototyping emphasizes exploring possibilities, extremes, and feasibility, providing essential references for final design choices. During this process, industrial designers actively and frequently make configurations and adjustments to prototype models, driven by their unpredictable intuitions, insights, and inspirations. Limited by cognitive bandwidth~\citep{lieder2020resource,griffiths2020understanding}, industrial designers expect immediate feedback on how their interventions affect the prototype's appearance without the need to switch attention to refining the current version of design~\citep{monsell2003task}, such as manually tuning the parameters of the \ac{cad} model through the interactive \ac{gui} of modeling engines. Such interruptions may disturb their continuous flow of thinking and creation~\citep{cross2023design}. Consequently, industrial designers have been on the quest of an \emph{interface} for the \emph{targeted} control of prototype models---one that seamlessly implements their intentions as though the modeling controller were an extension of their clay-playing hands, mirroring their thoughts, as shown in \cref{fig:motivation}. By providing simple \ac{nl}-based instructions, the interface should ensure the appropriate realization of their intentions, avoiding under-specification, where interventions are inadequately detailed, and over-specification, where unnecessary details distort the desired outcome.

Recent advancements in \ac{ai} have positioned \acp{llm} as a promising medium for interactions between industrial designers and modeling engines. These models can flexibly interpret designers' intentions through \ac{nl}-based instructions, comprehend and reason over the instructions, and generate corresponding modeling commands. Researchers have begun exploring \ac{llm}-based tools to support \ac{cad} model creation and editing~\citep{wu2023cad,yuan2024openecad}. However, the potential of \acp{llm} for fast prototyping with its targeted control demands has yet to be fully elicited, particularly in the context of the targetedness-demand fast prototyping~\citep{uusitalo2024clay}. In the practice of fast prototyping via these tools, the designers' intuitive, subjective, and convention-driven instructions for property assignment and adjustment are often not comprehended precisely, thus the prototype models may not be targetedly controlled. These drawbacks stem from the substantial gaps between the designers' languages for describing their intention, and the underlying programming languages that drives the modeling process through modeling commands. 

Designers' languages exhibit three key characteristics: they are high-level, ambiguous, and domain-specific. First, these languages employ semantic references to describe both component parts and their relationships. For example, in teapot design, designers refer to specific parts like ``pot lid'' and ``spout'', as well as relationships such as the ``relative position of pot lid and spout.'' This semantic approach ensures the language remains intuitive for designers' practical use~\citep{hannah2002elements}. Second, these languages inherently contain ambiguity~\citep{russell1923vagueness}. Consider how the term ``length'' carries different meanings when applied to different parts---the ``length'' of a ``pot lid'' refers to a distinctly different physical property than the ``length'' of a ``spout''~\citep{pei2011taxonomic}. This ambiguity extends to subjective operational instructions, such as ``make the shape of the spout sharper'', where the intended degree of modification remains uncertain. Third, these languages demonstrate strong domain-specificity. Each product category employs its own semantic vocabulary. Teapot design uses terms like ``pot lid'', ``spout'', and relationships like ``articulated'' and ``fused''. In contrast, wardrobe design employs different terminology such as ``base plate'', ``column'', ``inner'', and ``beneath''. This semantic tailoring extends across all product categories~\citep{micheli2012perceptions}. Following design convention, we use \emph{category} to denote products sharing similar functionality, while adopting the term \emph{domain-specificity} from knowledge representation theory to describe this category-specific variation in semantics~\citep{barsalou2008grounded}---a convention we maintain throughout this paper. In essence, designers' languages reflect human commonsense knowledge and intuitive thinking patterns. They approach product design through a top-down methodology, breaking down products into semantic components and their interrelationships, while considering their physical properties. This approach aligns with designers' natural cognitive processes and practical needs. 

In contrast to designers' languages, programming languages of modeling engines are low-level, precise, and domain-agnostic\footnote{The readers may refer to a commonly used, open-source \ac{cad} modeling language, FreeCAD. Visit the website at \url{https://www.freecad.org/}}. First, the atomic constructs of modeling commands---both their concepts and associated operations---contain no abstract meanings beyond basic graphical semantics~\citep{wilkinson2012grammar}. These commands operate solely on fundamental geometric elements such as points, curves, surfaces, shapes, intersections, and angles. This focus on basic geometric primitives ensures the commands maintain sufficient expressiveness for model construction. Second, modeling commands employ precise quantization and parameterization, diverging from the ambiguous descriptive adjectives common in natural language, such as ``big'' or ``sharp''. They also avoid relative comparatives like ``larger'' or ``sharper'' that humans typically use when assigning or adjusting properties~\citep{kennedy2007vagueness}. Third, the absence of abstract semantics in modeling commands renders them domain-agnostic rather than domain-specific in industrial design applications~\citep{mernik2005and}. While some encapsulated functions may provide abstract operations, such as batch processing similar parts, these functions remain bound by atomic operations and reflect the modeling engine's rendering mechanisms rather than addressing domain-specific design requirements~\citep{shi2023complexity}. In essence, modeling languages follow the logic of graphics description languages and avoid maintaining prefabricated components for specific requirements. Instead, they require employing a bottom-up approach and building meaningful semantic functions through the composition of atomic elements.

\begin{figure*}[t!]
    \centering
    \includegraphics[width=\linewidth]{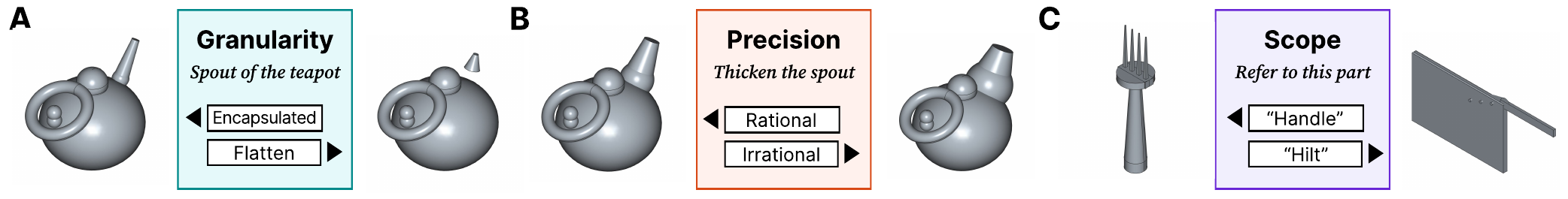}
    \vspace{-\baselineskip}
    \caption{\textbf{Illustrations of the gaps against the targeted control.} \textbf{(A)} Designers' language contains semantically rich, encapsulated concepts. Improper handling of abstraction levels can lead to undesired semantic flattening. \textbf{(B)} Designers' language relies heavily on subjective references that require commonsense knowledge for proper interpretation. Improper specification of instructions may result in physically or functionally inappropriate operations. \textbf{(C)} Designers' language is inherently domain-specific, where similar physical properties may be referenced by different semantic identifiers across product design domains.}
    \label{fig:empirical}
\end{figure*}

To bridge the gaps between designers' languages and modeling languages, we propose an interface built upon \acp{llm} that translates designers' high-level, ambiguous, and domain-specific intentions into low-level, precise, and domain-agnostic modeling commands. Our design principles emerge from a systematic study of fast prototyping practices and their cognitive foundations. The first principle establishes the interface as an intermediate-level \ac{dsl} between the two languages~\citep{fowler2010domain}. This \ac{dsl} uses designers' concepts and relationships as semantic identifiers, with each language construct encapsulating the atomic modeling commands needed to render parts or execute operations based on context. The abstraction level of this \ac{dsl} must carefully balance between the finest granularity of designers' languages and the broadest semantics of modeling languages~\citep{abelson1996structure}. The second principle addresses targeted prototype control through precise interpretation of designers' intentions. The interface \ac{dsl} must bind concepts (including parts and relationships) to their permissible operations. For instance, the ``length'' of a ``spout'' permits operations like ``increase'' or ``decrease,'' but not ``twist'' or ``smooth''---establishing a binding between a concept and its allowable second-order adjustments. Similarly, the ``relative position of pot lid and spout'' can be specified through directional descriptions, creating a binding between a relationship and its permissible first-order assignments~\citep{buring2005binding}. These bindings ensure the interface remains both expressive and intuitive~\citep{shi2024abstract}. The third principle acknowledges the domain-specific nature of the interface. While the \ac{dsl} offers benefits of hierarchical encapsulation and precise knowledge representation, each \ac{dsl} instance serves only one product design domain, as concepts and operations vary significantly across domains. Manual creation of domain-specific \acp{dsl} requires substantial effort and becomes impractical for design agencies given the diversity of consumer products~\citep{shi2024autodsl}. Therefore, to ensure practical utility, we propose an automated mechanism for domain-specific interface specification.

This study identifies the gaps between designers' and modeling languages in fast prototyping and proposes an interface to bridge these gaps. Our work makes three primary contributions: (i) we conduct a systematic analysis of the gaps between designers' languages and modeling languages in practice (\cref{sec:gaps}), deriving the design principles for the interface grounded in cognitive theory; (ii) we develop the interface architecture, detail its operational mechanisms, and design an algorithm for automated domain specification (\cref{sec:interface}); (iii) we evaluate the interface through both machine-based assessments and human studies (\cref{sec:result}), demonstrating its potential to bridge communication gaps between diverse-minded experts while serving as an auxiliary module for \ac{llm}-based interaction tools.

\section{The gaps against the targeted control}\label{sec:gaps}

In this section, we systematically study the practices of fast prototyping in industrial design and analyze the three gaps between the designers' and modeling languages, within the three dimensions: level of abstraction (\cref{subsec:gaps-abstraction}), semantic precision (\cref{subsec:gaps-precision}), and domain-specificity (\cref{subsec:gaps-domain}). According to these investigations, we propose the corresponding design principles of the interface (\cref{subsec:gaps-principle}).

\subsection{Level of abstraction and granularity}\label{subsec:gaps-abstraction}

This gap pertains to the level of abstraction and granularity of the language. In semantic theory, this aspect critically  influences the conceptual density and operational granularity of language~\citep{levinson2003space}. Designers employ a high-level, semantically rich language that encapsulates complex concepts and relationships inherent to specific design contexts. This language is closer to human natural language, which is inherently high-level and contextually enriched. On the other hand, modeling commands utilize a low-level, syntactically oriented language primarily focused on the precise execution of graphical operations. These commands operate at a lower level of abstraction, dealing directly with the geometrical and mathematical representations required to render objects, without the semantic enrichment seen in designers' language. As shown at \cref{fig:empirical}A, the designer's semantic \emph{``spout of teapot''} essentially encapsulates the modeling descriptions \emph{``a hollow cylinder thick at the proximal and thin at the distal''}.

\subsection{Semantic precision and determinacy}\label{subsec:gaps-precision}

This gap refers to the semantic precision and determinacy of the language. In the realm of semantics, this aspect is critical as it affects the interpretability and operational specificity of the language~\citep{lyons1995linguistic}. Designers' language exhibits a higher degree of semantic indeterminacy, allowing for contextual and subjective interpretation and flexibility. This ambiguity is typical in natural language, where meanings can shift based on context, usage, and individual perception. In contrast, modeling commands demonstrate a semantic determinacy, with each command having a fixed and unequivocal meaning, devoid of contextual interpretation. This precision is necessary for the accurate execution of graphical tasks, where ambiguity could lead to errors in design implementation. As shown at \cref{fig:empirical}B, the designer's intention of \emph{``thicken the spout''} implies a rational grounding of the diameter-related parameters, which can be irrational if not properly specified.

\subsection{Lexical scope and semantic applicability}\label{subsec:gaps-domain}

This gap highlights the lexical scope and semantic applicability of the language. From a semantic perspective, this involves the domain specificity or generality of the language~\citep{saeed2015semantics}. Designers' language is characterized by domain-specific lexicons, tailored to particular design fields or product categories. This specialization enables designers to communicate effectively within their specific context, using terminology and concepts that are relevant to their particular field. Conversely, modeling commands are domain-agnostic, designed to be universally applicable across various design fields without modification. This universality ensures that modeling engines remain versatile and broadly applicable, at the cost of not providing specialized tools that cater to the nuances of specific design domains. As shown at \cref{fig:empirical}C, referring to parts with similar physical properties within different product design domains require distinct semantic identifiers, such as \emph{``handle'' in ``fork''} and \emph{``knit'' in ``cleaver knife''}.

\subsection{Design principles for the interface}\label{subsec:gaps-principle}

The three gaps highlight fundamental differences in design conceptualization and communication. While designers' language is semantically rich, contextually flexible, and specialized, modeling commands are syntactically precise, context-independent, and generalized. To bridge these gaps, we propose three design principles derived from the tacit knowledge of industrial design practices.

\paragraph{Balanced hierarchy of abstraction}

This principle addresses the alignment between designers' semantic language and modeling systems' rigid syntax through an intermediate-level \ac{dsl}. This \ac{dsl} is designed to preserve \emph{naturalness} by maintaining the conceptual richness of designers’ language~\citep{myers2004natural,hindle2016naturalness}, such as terms like ``organic form'' or ``balanced proportions'', while abstracting low-level geometric operations like extrusion tolerances or fillet radii into executable functions. Crucially, the interface ensures that granular modeling details remain transparent to designers, allowing them to focus on intent rather than implementation. For instance, an instruction to \emph{``soften the edges for an ergonomic feel''} maps to parametric functions adjusting fillet radii without exposing the underlying modeling commands. To achieve this balance, the \ac{dsl} employs \emph{function-like natural language constructs} that encapsulate atomic graphical elements into reusable semantic identifiers~\citep{bryant2011computer,gulwani2017program}. This is achieved through a reciprocal development process: (i) top-down decomposition of designers’ domain-specific terminology (\eg, breaking ``part handle'' into geometric primitives like ``angled cylinders with blended intersections'') establishes a semantic anchor for translation; while (ii) bottom-up encapsulation aggregates atomic modeling commands (\eg, Boolean union, surface lofting) into higher-order functions (\eg, ``generate grip geometry'') that mirror designers’ mental models. By harmonizing these bidirectional strategies, the \ac{dsl} operates at a “meso-scale” abstraction—neither oversimplifying creative intent nor overcomplicating execution—thereby sustaining designers’ cognitive flow while guaranteeing technical precision.

\paragraph{Articulated concepts and operations}

This principle ensures precise translation of designers' intentions through concept-operation bindings that reflect domain-specific design logic. This is achieved by binding concepts (\ie, geometric parts or relational configurations) to spaces of permissible operations that reflect domain-specific design logic. These bindings enforce semantic coherence by restricting operations to contextually valid transformations, thereby preventing mismatches between intent and execution. Hence, the interface employs a \emph{hierarchical syntax} where operations are linked to governing concepts~\citep{chomsky1957syntactic,shi2024expert}. The effective field of an operation is confined to the attributes of the associated concept. First-order operations (\eg, static descriptors like ``smooth'' or ``short'') directly modify attribute values, while second-order operations (\eg, dynamic comparatives like ``smoother'' or ``shorter'') adjust the incremental magnitude of these modifications~\citep{emde1983discovery}. Critically, the system maintains spatial-temporal continuity~\citep{kuipers1986qualitative}: attributes remain stable until explicitly modified, and operational effects stay consistent until recalibrated. This structure balances expressive power with cognitive clarity~\citep{felleisen1991expressive}, enabling precise intent articulation while avoiding excessive modeling choices.

\paragraph{Automated domain specification}

This principle confronts the practical challenge of scaling the interface’s utility across diverse product design domains, where manual crafting of \acp{dsl} becomes prohibitively costly due to the vast heterogeneity of concepts and operations~\citep{pei2011taxonomic}. The interface integrates a mechanism that autonomously derives domain-specific linguistic and operational landscapes by modeling the joint distribution of designers’ language and modeling command patterns. Drawing from cognitive theories of \emph{concept acquisition}~\citep{kemp2008discovery}, where humans construct mental models by iteratively sampling and generalizing from exemplars~\citep{tenenbaum2011grow}, the mechanism leverages \acp{llm} as repositories of commonsense knowledge to build generative models of design concepts~\citep{yildirim2024task}. By framing \ac{llm} interactions as a stochastic sampling process, the system emulates human-like concept formation, where repeated exposure to diverse examples, mediated by the \ac{llm}’s latent knowledge, gradually defines the boundaries and operational rules of domain-specific concepts. This approach not only captures the breadth of real-world design variation but also ensures that the synthesized \ac{dsl} remains grounded in both technical feasibility and designers’ linguistic norms. This approach transforms manual specification into a scalable, knowledge-driven workflow~\citep{shi2024autodsl,shi2025hierarchically}, ensuring comprehensive coverage while maintaining practicability.

\begin{figure*}[t!]
    \centering
    \includegraphics[width=\linewidth]{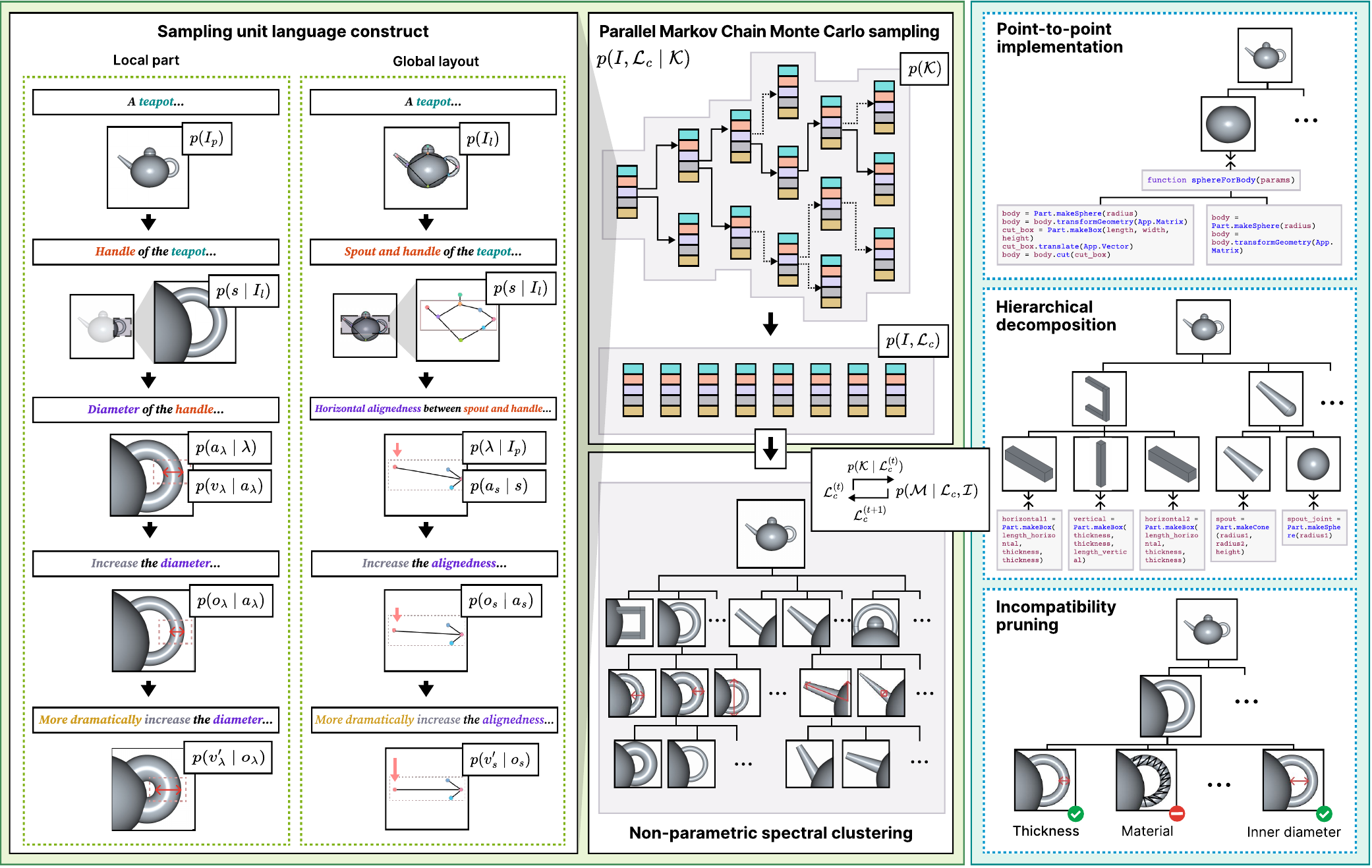}
    \vspace{-\baselineskip}
    \caption{\textbf{Representation and domain adaptation of the interface.} This illustration conveys a running example of adapting the interface to the teapot design domain. The iterative optimization alternates \textbf{\textcolor{highlight-e-step}{construct expansion (left)}} with \textbf{\textcolor{highlight-m-step}{feasibility validation (right)}}.}
    \label{fig:framework}
\end{figure*}

\section{The interface for the targeted control}\label{sec:interface}

In this section, we describe our proposed methodology for the target control of fast prototyping. We first present the problem formulation (\cref{subsec:interface-problem}), followed by the interface representation (\cref{subsec:interface-represent}), and we describe the algorithm for the domain adaptation of the interface (\cref{subsec:interface-learn}; \cref{fig:framework}).

\subsection{The intention communication problem}\label{subsec:interface-problem}

\paragraph{Original formulation}

We consider the original problem of communicating the designers' intention with the modeling engine as a mapping $f:\mathcal{I}_c\mapsto\mathcal{M}$ that transforms the space of intention $\mathcal{I}_c$ to the space spanned by the possible command constructs of the modeling engine $\mathcal{M}$. Each instance of intention $I\in\mathcal{I}_c=\{I_l,I_p\}$ is translated to a sequence of modeling commands $M\in\mathcal{M}=\langle m_1,\dots,m_{|M|}\rangle$, where $I_l$ represents the intention of configuration on the global shape, structure, and layout of the set of parts of the target product, and $I_p$ denotes the intention of configuration on the parts of the target product; $m_k$ represents a program indicating a modeling command. Hence, the objective of modeling according to the designers' intention can be maximizing the likelihood
\begin{equation}
    \arg\max_{M}\;p(M\mid I).
\end{equation}

\paragraph{Integrating interface into the formulation} 

Directly estimating the likelihood $p(M\mid I)$ can be intractable, given the substantial gaps between the designers' and modeling languages. We introduce the intermediate-level \ac{dsl}, the interface $\mathcal{L}_c$ into the problem, decomposing the original mapping into a combination of mappings $g:\mathcal{I}_c\mapsto\mathcal{L}_c$ and $h:\mathcal{L}_c\mapsto\mathcal{M}$, where $f=h\circ g$. The generative process can then be formulated as $\mathcal{I}_c\xrightarrow{g} \mathcal{L}_c\xrightarrow{h} \mathcal{M}$. The interface \ac{dsl} $\mathcal{L}_c=\{\mathcal{S}_c,\Lambda_c\}$, where $\mathcal{S}_c$ represents the set of \ac{dsl} constructs regarding the global shape, structure, and layout of the set of parts of the target product, and $\Lambda_c$ denotes the set of \ac{dsl} constructs regarding the parts of the target product. An arbitrary permissible program generated by the \ac{dsl} can thereby be represented as $L\in \mathcal{L}_c^*=\{\text{prog}\mid \text{prog}\in\mathcal{S}_c^*\cup\Lambda_c^*\}$. Treating the interface as a latent variable in the likelihood maximization objective, we have
\begin{equation}
    \begin{aligned}
        &\arg\max\limits_{M} p(M\mid I) \\
         =&\arg\max\limits_{L}\sum\limits_{L\in\mathcal{L}_c^*}p(M\mid L)p(L\mid I),
    \end{aligned}
\end{equation}
where $p(L\mid I)$ and indicates $p(M\mid L)$ generative functions $g$ and $h$ respectively. Leveraging this latent variable formulation, we obtain a relatively tractable estimation of $p(M\mid I)$ with the interface as the medium.

\subsection{Representation of the interface}\label{subsec:interface-represent}

\paragraph{Syntax of the \ac{dsl}}

The local part construct is defined as $\Lambda_c=\langle\lambda,a_{\lambda},o_{\lambda},v_{\lambda},v_{\lambda}'\rangle$, where $\lambda$ denotes the semantic identifier of the referred part, $a_{\lambda}=(a_{\lambda,i},a_{\lambda,p})$ indicates the physical subpart $a_{\lambda,i}$ (\eg, a handle or its constituent cylinder) and its associated physical property $a_{\lambda,p}$ (\eg, length, diameter), $o_{\lambda}$ represents the operation that is permissible to apply on the physical property, $v_{\lambda}$ denotes the quantitative value of the physical property (\ie, first-order value assignment), and $v_{\lambda}'$ indicates the quantitative value of the extent of the operation (\eg, second-order value adjustment). The global construct is defined as $\mathcal{S}_c=\langle s,a_s,o_s,v_s,v_s'\rangle$, where $s$ denotes the semantic identifier of the referred relationship between parts, $a_s$ indicates the joint descriptor of the physical subparts within the two parts (\eg, the relative position of spout and pot lid), $o_s$ represents the combinatorial operation over the relationship (\eg, make the relative position closer), $o_s$ represents the combinatorial operation applied to the relationship, and $v_s'$ combines the values of the two operations linked to the physical properties. The syntax of these language constructs mirrors the designer's intention, with each unit maintaining an identical structural pattern.

\paragraph{Hierarchical modeling}

According to our semantic binding design principle, we describe the \ac{dsl} that generates language constructs using a top-down approach. In this hierarchy, lower-level elements are controlled by their immediate higher-level elements: parts are generated based on the intention, physical subparts are generated from the part, operations are determined by physical properties, quantitative values are assigned according to their corresponding physical properties and operations. To maintain \ac{dsl} expressivity that accommodates the high variability in designers' intentions for the same target product~\citep{shi2024constraint}, we model the relationships between adjacent levels as conditional probabilistic generation~\citep{wu2019tale}. This results in a tree-structured hierarchical probabilistic model, which can be formulated as
\begin{equation}
    \begin{aligned}    
    &p(I,\mathcal{L}_c)\\=&p(I_l,I_p,s,a_s,o_s,v_s,v_s',\lambda,a_{\lambda},o_{\lambda},v_{\lambda},v_{\lambda}')\\=&p(v_{\lambda}'\mid o_{\lambda})p(v_s'\mid o_s)p(v_{\lambda}\mid a_{\lambda})p(o_{\lambda}\mid a_{\lambda})\\ &p(v_s\mid a_s)p(o_s\mid a_s)p(a_{\lambda},a_s\mid\lambda, s)p(s,\lambda\mid I_l, I_p),
    \end{aligned}
\end{equation}
representing the overall joint distribution of the designers' language and the interface.

\subsection{Domain adaptation of the interface}\label{subsec:interface-learn}

\begin{figure}[t!]
    \centering
    \includegraphics[width=\linewidth]{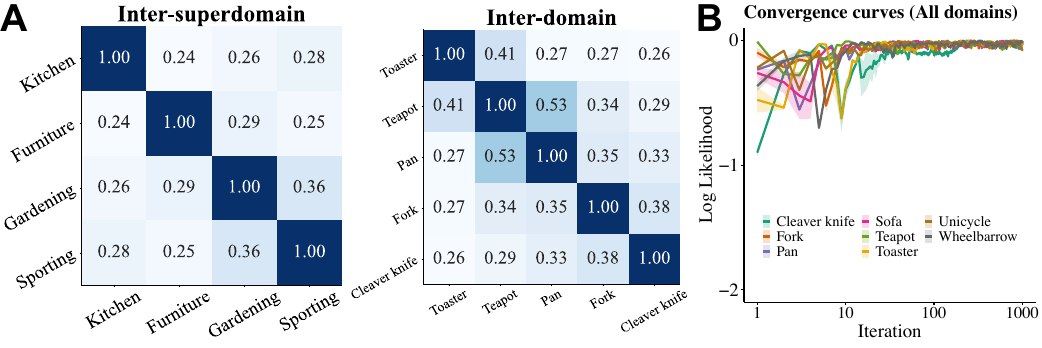}
    \vspace{-\baselineskip}
    \caption{\textbf{Running status of the domain adaptation algorithm.} \textbf{(A)} Distinctions on language construct distributions cross supercategories and domains. \textbf{(B)} Convergence curves of the domain adaptation algorithm on eight domains.}
    \label{fig:running}
\end{figure}

The target of domain adaptation is to automate the design of the interface \ac{dsl} $\mathcal{L}_c$ given any domain of design $c$ (\eg, teapot, sofa, fence, racket, \etc). According to our design principle of domain adaptation, we regard interacting with \acp{llm} as a Bayesian inference from the commonsense knowledge base $\mathcal{K}$ with prior distribution $p(\mathcal{K})$, serving as the prior of a latent generative model
\begin{equation}
    p(I,\mathcal{L}_c,\mathcal{K})=p(I,\mathcal{L}_c\mid\mathcal{K})p(\mathcal{K}),
\end{equation}
where $\mathcal{L}_c$ is iteratively refined through stochastic sampling. To achieve this objective, wherein the interface \ac{dsl} must accept the designers' language and generate the modeling language, we derive the iterative algorithm with a reciprocative optimization framework (see \cref{fig:framework,fig:running}).

\paragraph{Sampling from commonsense priors}

To initialize $\mathcal{L}_c$, we sample language constructs $\Lambda_c$ and $\mathcal{S}_c$ from the \ac{llm}-mediated prior $p(\mathcal{K})$, following their hierarchical architecture level-by-level. This is implemented as a \ac{mcmc} process. The \ac{mcmc} is initialized by drawing $M$ seed sample constructs $\{L_1^{(0)},\dots,L_M^{(0)}\}$ via \ac{llm} queries. For each seed sample $L_i^{(0)}$, $N$ Metropolis-Hastings steps are performed to explore the neighborhood of $\mathcal{K}$, specified as
\begin{equation}
    L^{(t+1)}\sim q(L'\mid L^{(t)})\min\bigg(1,\frac{p(L'\mid\mathcal{K})}{p(L^{(t)}\mid\mathcal{K})}\bigg),
\end{equation}
where the proposal distribution $q$ is defined by \ac{llm}-generated perturbations. This process yields $M\times N$ samples $\{L_i^{(t)}\}$, ensuring diversity while preserving domain coherence of the samples. The sampled constructs are clustered into semantically coherent units exploiting a non-parametric \ac{dpmm} to handle the unknown and variable number of domain concepts. Let $\theta_k$ denote the parameters of the $k$-th cluster (\eg, one type of the handle within teapot), with a Chinese Restaurant Process prior~\citep{blei2010nested}, each sample is assigned to cluster $k$ with probability proportional to its likelihood under $p(L_i^{(t)}\mid \theta_k)$ modeled as a multinomial distribution over the elements $\langle\lambda,a_{\lambda},o_{\lambda},\dots\rangle$ and $\langle s,a_s,o_s,\dots\rangle$. The \ac{dpmm} induces a hierarchical tree structure, where nodes closer to the root represent macro concepts (\eg, part), and nodes closer the leaves encode fine-grained variants (\eg, physical properties).

\paragraph{Reciprocative optimization of granularity}

Though the sampling drawing and spectral clustering processes constructs the initial structure of the interface, there is one side missing---the interface not only accepts the designers' language, but also outputs the modeling language. The actual granularity of the interface \ac{dsl} is determined by the both sides. Therefore, we exploit an \ac{em} Algorithm-like reciprocative optimization that alternates between \emph{construct expansion} (\ie, Expectation) and \emph{feasibility validation} (\ie, Maximization). The Expectation step expands the \ac{dpmm} tree structure by sampling new constructs and proposing new clusters using the current interface $\mathcal{L}_c^{(t)}$ as a prior. The Maximization validates the constructs of the interface \ac{dsl} via simulated synthesize over the programming library of the modeling language. The validation phase categorizes each construct into three feasibility classes: (i) \emph{point-to-point implementation}, which can be implemented by atomic operations or predefined functions within the modeling language, then the construct should be maintained; (ii) \emph{hierarchical decomposition}, which may be implemented by some encapsulated functions within the modeling language, the construct sample should be regenerated at a finer granularity; (iii) \emph{incompatibility pruning}, which cannot be implemented by the modeling language, such as the physical property ``material'', the construct should be removed. After this Maximization-step, the algorithm goes into the next iteration. The likelihood is maximized by iterating
\begin{equation}\label{eq:m-step}
    \mathcal{L}_c^{(t+1)}=\arg\max_{\mathcal{L}_c^{(t)}}\mathbb{E}_{p(\mathcal{K}\mid\mathcal{L}_c^{(t)})}[\log p(\mathcal{M}\mid\mathcal{L}_c,\mathcal{I})],
\end{equation}
where $p(\mathcal{K}\mid\mathcal{L}_c^{(t)})$ indicates the mechanism to focus exploration of $\mathcal{K}$ based on the interface’s current state. Intuitively, this mechanism indicates that given the current interface state, what areas of the commonsense knowledge are most relevant to refine or expand it, mirroring human designers’ iterative refinement---prior knowledge is recalled in context of current progress.

\begin{figure*}[t!]
    \centering
    \includegraphics[width=\linewidth]{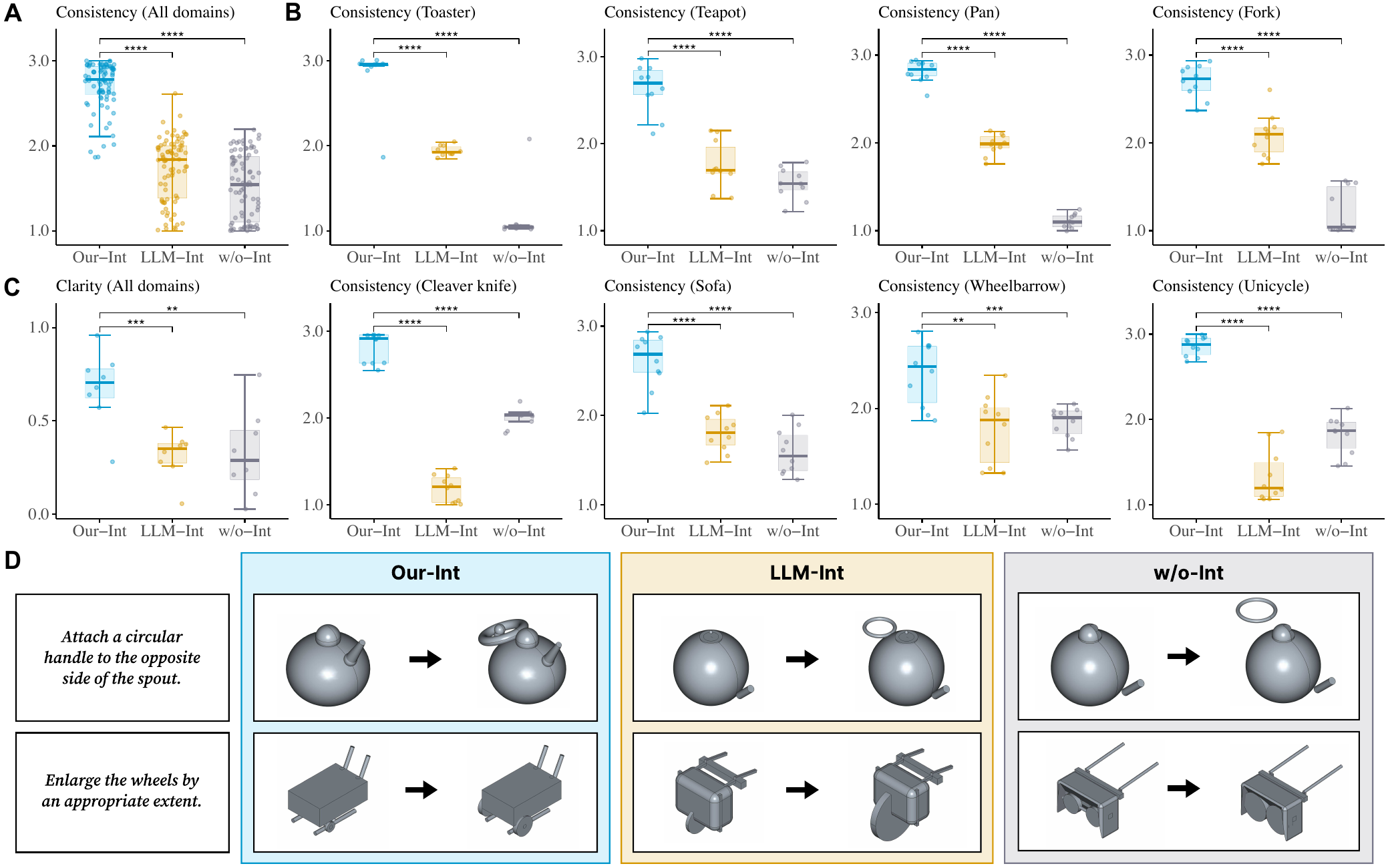}
    \vspace{-\baselineskip}
    \caption{\textbf{Quantitative and qualitative results.} \textbf{(A)} Results on rendering consistency assessment across all domains. \textbf{(B)} Isolated domain-wise results on rendering consistency assessment. \textbf{(C)} Results on information clarity assessment across all domains. \textbf{(D)} Showcases collected from the fast prototyping tasks. The leftmost text is the designer's instruction. Each pair of visualizations indicates the models before (left to the arrow) and after (right to the arrow) the intervention.}
    \label{fig:results}
\end{figure*}

\section{Experiments and results}\label{sec:result}

In this section, we report and discuss the results of our experiments for assessing the utility of the interface for targeted control. We start from describing our designer-participated fast prototyping tasks (\cref{subsec:result-setup}), along with the metrics to measure the targetedness of the control of designers' intentions on the modeling engine (\cref{subsec:result-metrics}). Afterwards, we introduce the alternative methods equipping integrating different interfaces used for comparison (\cref{subsec:result-methods}). Finally, we report and analyze the experimental results both quantitatively and qualitatively (\cref{subsec:result-res}).  

\subsection{Experimental setups}\label{subsec:result-setup}

\paragraph{Experimental protocol}

We evaluate the targetedness of modeling control across two dimensions: rendering consistency and information clarity. We design a realistic fast prototyping scenario to simulate the targetedness-demanding design practices. Given the fundamental differences between fast prototyping and production-ready modeling~\citep{hallgrimsson2012prototyping}, we limit the iteration count to ten rather than allowing unlimited iterations until the modeling becomes subjectively ``satisfying''. At the beginning, participants receive information about their target domain of product design. During each iteration, participants provide one natural language instruction to the interface and receive images of rendered models generated by our interface and alternative methods. Participants then rank these images based on how closely they match their intended effect. We collect these rankings for consistency measurement. Additionally, we gather the modeling programs generated during this process for machine-based evaluation.

\paragraph{Participants}

The human study includes 50 participants, each holding at minimum a Bachelor's degree in industrial design or related fields. All participants receive standardized task instructions and provide informed consent. For ethical considerations, please refer to \cref{sec:supp-ethics}.

\paragraph{Materials}

To ensure experimental generality, we select eight product design targets from major consumer market supercategories~\citep{pei2011taxonomic}: \emph{kitchen appliances} (\ie, teapot, pan, fork, toaster, and cleaver knife), \emph{furniture} (\ie, sofa), \emph{gardening equipment} (\ie, wheelbarrow), and \emph{sporting goods} (\ie, unicycle). We choose products across these supercategories to examine inter-supercategory differences and selected multiple products within supercategories to study intra-supercategory domain-wise variations. To thoroughly test interface utility, we intentionally select challenging products with irregular geometric features, such as teapots, sofas, wheelbarrows, and unicycles.

\subsection{Evaluation metrics}\label{subsec:result-metrics}

\paragraph{Rendering consistency assessment}

We measure the consistency between participant intentions and modeling results using the ranked order of results from each intervention step, following established methodology in similar human studies~\citep{lake2015human}. We treat the iterative process as a Markov Process, considering only transitions between adjacent design versions, as each intervention represents a state transition. Therefore, we apply uniform weighting to step-wise result rankings across each design task, with higher scores indicating better performance.

\paragraph{Information clarity assessment}

For each participant intervention step, we analyze the corresponding modeling programs for uncertainty by examining the modeling program library documentation~\citep{chaitin1977algorithmic}. We analyze function \acp{ast} and calculate uncertainty based on the cumulative depth of specified program constructs along the resulting \ac{ast}. Greater cumulative depth indicates lower uncertainty and higher information clarity. Higher scores represent better performance.

\subsection{Alternative methods}\label{subsec:result-methods}

Our comparison with alternative methods are required to answer two key questions. First, is our interface architecture necessary for the targeted control, specifically the transformation from standard likelihood maximization to our latent variable interface approach? Second, are our interface implementation principles, particularly the hierarchical \ac{dsl}-based representation, essential for the targeted control? To address the first question, we develop \texttt{w/o-Int}, an alternative model that processes designers' language directly without interfaces. For the second question, we created \texttt{LLM-Int}, an interface implemented through pure \ac{llm}-based prompt engineering, as a direct comparison to our proposed interface. We refer to our \ac{dsl}-represented interface model as \texttt{Our-Int}. Each domain-specific instance of \texttt{Our-Int} is automatically designed through the algorithm described in \cref{subsec:interface-learn}. All methods are built upon state-of-the-art text-to-modeling program generators~\citep{wu2023cad,yuan2024openecad}.

\subsection{Results}\label{subsec:result-res}

\paragraph{Rendering consistency}

Paired samples t-tests reveal that \texttt{Our-Int} significantly outperforms both alternative methods across all eight domains in rendering consistency (\texttt{Our-Int} \vs \texttt{LLM-Int}: $t(7183)=80.60,p<.0001$; \texttt{Our-Int} \vs \texttt{w/o-Int}: $t(7246)=63.81,p<.0001$; see \cref{fig:results}A)\footnote{The varying degrees of freedom in different t-tests result from some participants only selecting results with highest consistency without completing the full ranking. Such incomplete data cases represent less than $2\%$ of all collected data.}. Domain-specific analyses also show significant superiority of our approach in isolated comparisons (see \cref{fig:results}B). Furthermore, \texttt{LLM-Int} demonstrates significant overall improvement compared to \texttt{w/o-Int} ($t(7127)=12.37,p<.0001$; see \cref{fig:results}A), supporting the necessity of the interface structure. These findings validate our approach from a black-box outcome perspective.

\paragraph{Information clarity}

Paired samples t-tests demonstrate that \texttt{Our-Int} significantly outperforms both alternative methods across all eight domains in information clarity (\texttt{Our-Int} \vs \texttt{LLM-Int}: $t(14)=4.413,p<.001$; \texttt{Our-Int} \vs \texttt{w/o-Int}: $t(14)=3.277,p<.01$; see \cref{fig:results}C). These results validate our proposal from a white-box intermediate result perspective.

\paragraph{Discussion}

Comparison between \texttt{Our-Int} and alternative methods reveals significant advantages in targeted control: (i) while alternative methods only modify parameters at a coarse level, \texttt{Our-Int} successfully handles fine-grained instructions such as ``attach'' and ``opposite'' (see \cref{fig:results}D); (ii) \texttt{Our-Int} precisely interprets subjective references to values and operation extents, and also domain-specific references to parts and physical properties, while alternative models often reference incorrect objects for operations, apply inappropriate operation types, and generate physically implausible results (\eg, oversized wheels conflicting with the framework in the wheel size adjustment case; see \cref{fig:results}D). These qualitative observations further support our interface's utility. We notice that the increased complexity of generated modeling programs led to higher probabilities of run-time errors during modeling engine execution. However, this limitation is relatively minor, as current \ac{llm}-based program fixers can iteratively refine programs based on error messages~\citep{madaan2023self}.

\section{General discussion}

In this study, we propose an interface to enhance targeted control of fast prototyping by bridging the gap between designers' languages and modeling languages. Experimental results demonstrate our interface's effectiveness and suggest its potential as an auxiliary module for \acp{llm} in broader human instruction specification tasks.

\paragraph{On the cost of inter-domain communication}

Communication costs present a significant challenge in collaborative efforts. Domain experts from different backgrounds come with diverse mindsets, thereby using distinct \acp{dsl} accordingly, which often remain misaligned without extensive trial-and-error collaboration~\citep{shi2023perslearn}. This challenge is particularly evident in industrial design and manufacturing, where designers must interpret high-level Part-A indicators, requirements, and preferences; while modeling engineers and their tools must implement low-level designs within spatial, graphical, and physical constraints. The substantial language differences between these groups traditionally result in high communication costs. Our interface aims to reduce these costs, potentially enabling a more balanced and reciprocal relationship between design creation and prototype implementation. The broader prospect of our methodology is to provide human experts with a \emph{natural language programming} interface that enables direct high-level instructions, while low-level executions are encapsulated under semantic identifiers within the interface, remaining transparent to domain experts.

\paragraph{On the domain generalizability of the interface}

A key consideration is whether interface representations should be specific or general across product design domains. While a universal interface might seem appealing, creating a truly general \emph{one-size-fits-all} solution for all product design domains presents significant challenges. Although theoretically possible to develop a comprehensive designers' language covering all requirements, such a system would become overwhelmingly complex, with overlapping semantics and intertwined references that designers would find impractical. Conversely, attempts to simplify this universal interface would inevitably compromise its representational capabilities, leading to the expressivity-complexity dilemma~\citep{abelson1996structure}. Therefore, rather than pursuing elusive generality, a more pragmatic approach may be to focus on developing and automating the adaptation of domain-specific interfaces.

\section*{Acknowledgements}

This work is partially supported by the National Natural Science Foundation of China under Grants 52475001 and RGC GRF Grant 16210321. Q. Xu is a visiting student at Peking University from University of Science and Technology of China.

\section*{Impact statement}

This work explores bridging high-level industrial design instructions and low-level modeling implementations in a cost-efficient manner for fast prototyping of industrial designs through the automatic design and deployment of domain-specific interfaces. Our proposed methodology extends beyond industrial design to address a fundamental challenge in smart manufacturing: aligning requirements from design teams (Part-A) with production capabilities (Part-B). This alignment has long been a costly bottleneck that limits manufacturing efficiency, particularly given the current trend toward small-batch, multi-variety production. The challenge manifests in three critical gaps: level of abstraction, semantic precision, and domain-specificity, which together hinder rapid product iteration.

In response to these challenges, this work highlights the importance of managing tacit knowledge in manufacturing. The \emph{natural \ac{dsl}} used by specific Part-A or Part-B teams should be \emph{standardized toward a sound and complete \ac{dsl}} within their respective domains to avoid insufficient detail, ambiguity, and excessive jargon. Interfaces serve as information-preserving translators that map standardized Part-A \ac{dsl} programs to executable Part-B \ac{dsl} programs, enabling customized point-to-point communication between any pair of Part-A and Part-B components.

From a system-level perspective of the entire manufacturing supply chain, all suppliers are interconnected through a communication \emph{busline}. The interfaces provide suppliers with access to this busline while incorporating the unified requirements of the \ac{oem} that governs the supply chain. These interfaces can be customized for each supplier at low cost because they are designed through highly automated processes. This automation is practical given the substantial volume of historical communication data records between Part-A and Part-B components, combined with \acp{llm} for manufacturing that serve as commonsense knowledge bases. This standardization methodology has the potential to improve both the quality and efficiency of manufacturing supply chains.


\bibliographystyle{icml2025}
\bibliography{references}

\clearpage
\newpage
\onecolumn
\appendix

\renewcommand\thefigure{A\arabic{figure}}
\setcounter{figure}{0}
\renewcommand\thetable{A\arabic{table}}
\setcounter{table}{0}
\renewcommand\theequation{A\arabic{equation}}
\setcounter{equation}{0}
\pagenumbering{arabic}
\renewcommand*{\thepage}{A\arabic{page}}
\setcounter{footnote}{0}

\section{Additional remarks}\label{sec:supp-remarks}

\subsection{Rationale for the proposed methodology}\label{subsec:supp-remarks-dpmm}

Notably, this work is not an alternative to \ac{llm}-based \ac{cad} generators, but rather an interface to improve the performance of \ac{llm}-based \ac{cad} generators by bridging designers' language with modeling engineers' language. The fundamental challenge is that while modeling languages are hierarchically documented, designers' language is not standardized in the wild. This necessitates a systematic representation of concepts described in designers' language---from product categories to structures, components, attributes, and operations. This requirement aligns with word learning in cognitive science, where humans learn systems of interrelated concepts rather than isolated terms. 

Drawing inspiration from cognitive development, we mirror how people learn concept networks by sampling from the environment and organizing these samples into hierarchical structures through \ac{dpmm}. This spectral clustering approach captures multi-level attributes rather than clustering based on overall similarity. In our approach, we substitute environmental sampling with LLM-generated samples, as \acp{llm} are recognized repositories of commonsense knowledge. We then apply \ac{dpmm} to cluster these samples into a hierarchy of structures, components, attributes, and operations. This systematic representation allows natural instructions from designers to be decomposed into fine-grained elements that align with modeling language constructs, enabling targeted control in fast-prototyping.

\subsection{Rationale for the selected evaluation metrics}\label{subsec:supp-remarks-metric}

Fast prototyping allowing designers to explore brainstormed ideas without elaborating their instructions into modeling engineers' language. Our approach is explicitly two-stage, with our proposed interface serving as the first stage and \ac{llm}-based \ac{cad} generators as the second. Our metrics were selected to measure specific aspects of the process. Specifically, we use rendering consistency to measure how well the interface captures designers' intentions, essentially evaluating if desired elements appear and undesired ones are suppressed. Meanwhile, information clarity metrics quantify how effectively information transfers between designers' high-level language and modeling engineers' fine-grained requirements.

\subsection{Rationale for the protocol of design study}\label{subsec:supp-remarks-protocol}

Fast prototyping differs from full design. While full design requires master models for mass production, fast prototyping serves as an exploration for primary design ideas. Design studies often use two setups: (i) counting iterations to reach certain results; or (ii) evaluating improvements within a fixed iteration count. We adopt the latter to assess the improvements of each iteration with the same instruction, therefore relatively invariant to the number of iterations. The current choice of 10 is informed by discussions with professional industrial designers to capture the typical lifecycle of fast prototyping, and would be further investigated in further study.

Also, we would like to clarify that we aim to directly assess the targetedness of each individual design instructions, \eg, ``make the spout narrower''. This measure is somewhat subjective, as designers must evaluate whether desired changes were implemented while undesired changes were suppressed. Our work requires step-by-step assessment of each instruction within a sequence leading to a final product, whereas existing datasets typically evaluate only the final result after multiple instructions. This fundamental difference means that while conventional methods can create groundtruth references in advance, our approach relies on designers' on-the-fly instructions, making it impractical to prepare ground truth data (\eg, point clouds) beforehand. 

\subsection{Rationale for not comparing with more baselines}\label{subsec:supp-remarks-comparison}

We would like to first clarify that our work is not proposing an alternative to \ac{llm}-based \ac{cad} generators, but rather an interface to improve the performance of those generators. Our approach is explicitly two-stage, with our interface serving as the first stage and \ac{llm}-based \ac{cad} generators as the second. To our knowledge, there are no established state-of-the-art baselines for direct comparison of the entire two-stage pipeline. Comprehensive baselines exist for the second stage, so we have adopted the current state-of-the-art to evaluate the first stage, subject to our proposed interface.

\subsection{Relations with program synthesis}\label{subsec:supp-remarks-ps}

Our automated interface design shares conceptual similarities with program synthesis in achieving hierarchical abstraction through iterative sampling and refinement. However, two key distinctions exist. The first lies in the perspective of knowledge representation. Program synthesis relies on structured knowledge, using exemplar \ac{dsl} programs to generate higher-level libraries. In contrast, our approach samples from unstructured commonsense knowledge bases (\eg, \acp{llm}) and organizes knowledge into a \ac{dsl}. The second comes from the perspective of machine learning paradigm. Program synthesis follows a supervised approach, leveraging I/O pairs with task specifications, whereas our method is unsupervised, emerging from commonsense knowledge bases.

\section{Ethics statement}\label{sec:supp-ethics}

The human studies included in this work has been approved by the \ac{irb} of Peking University. We have been committed to upholding the highest ethical standards in conducting this study and ensuring the protection of the rights and welfare of all participants. We paid the participants a wage of \$22.5/h, which is significantly higher than the standard wage. 

We have obtained informed consent from all participants, including clear and comprehensive information about the purpose of the study, the procedures involved, the risks and benefits, and the right to withdraw at any time without penalty. Participants were also assured of the confidentiality of their information. Any personal data collected (including name, age, and gender) was handled in accordance with applicable laws and regulations.

\section{Experimenting details}\label{sec:supp-exp}

\subsection{Resulting models from fast prototyping}\label{subsec:supp-exp-model}

\begin{figure}[ht]
    \centering
    \includegraphics[width=\linewidth]{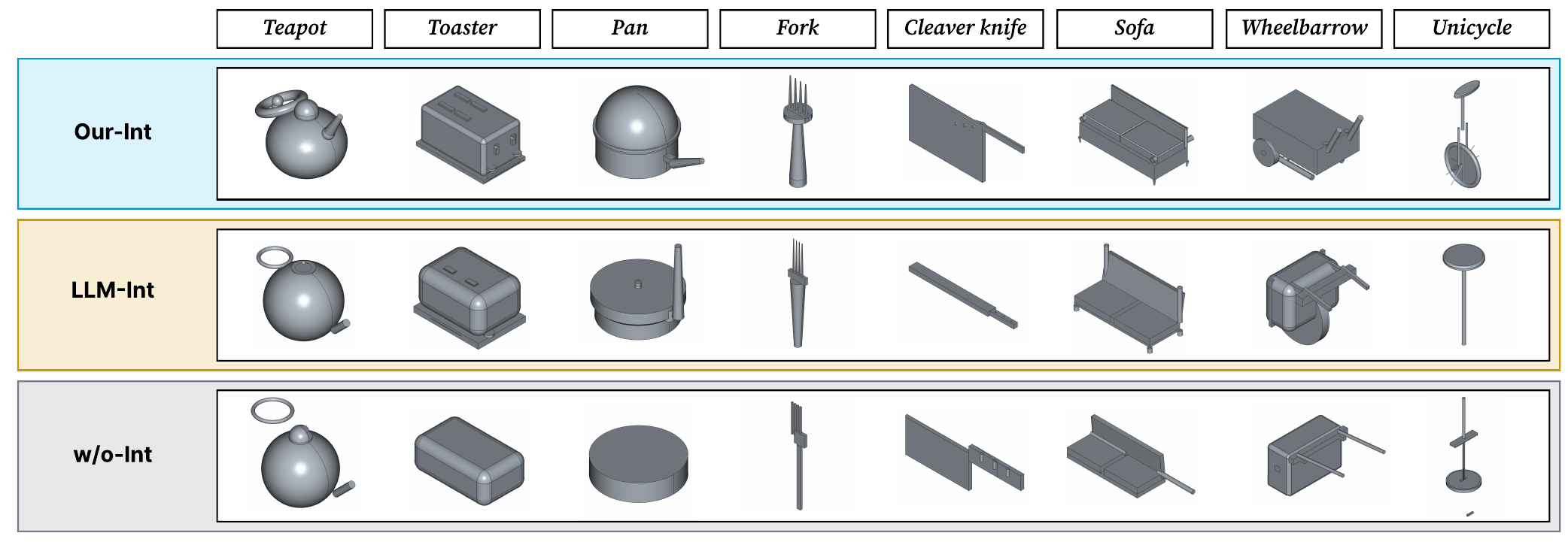}
    \vspace{-\baselineskip}
    \caption{\textbf{Visualization of the modeling results from the fast prototyping tasks}}
    \label{fig:supp-models}
\end{figure}

\begin{figure}[ht!]
    \centering
    \includegraphics[width=\linewidth]{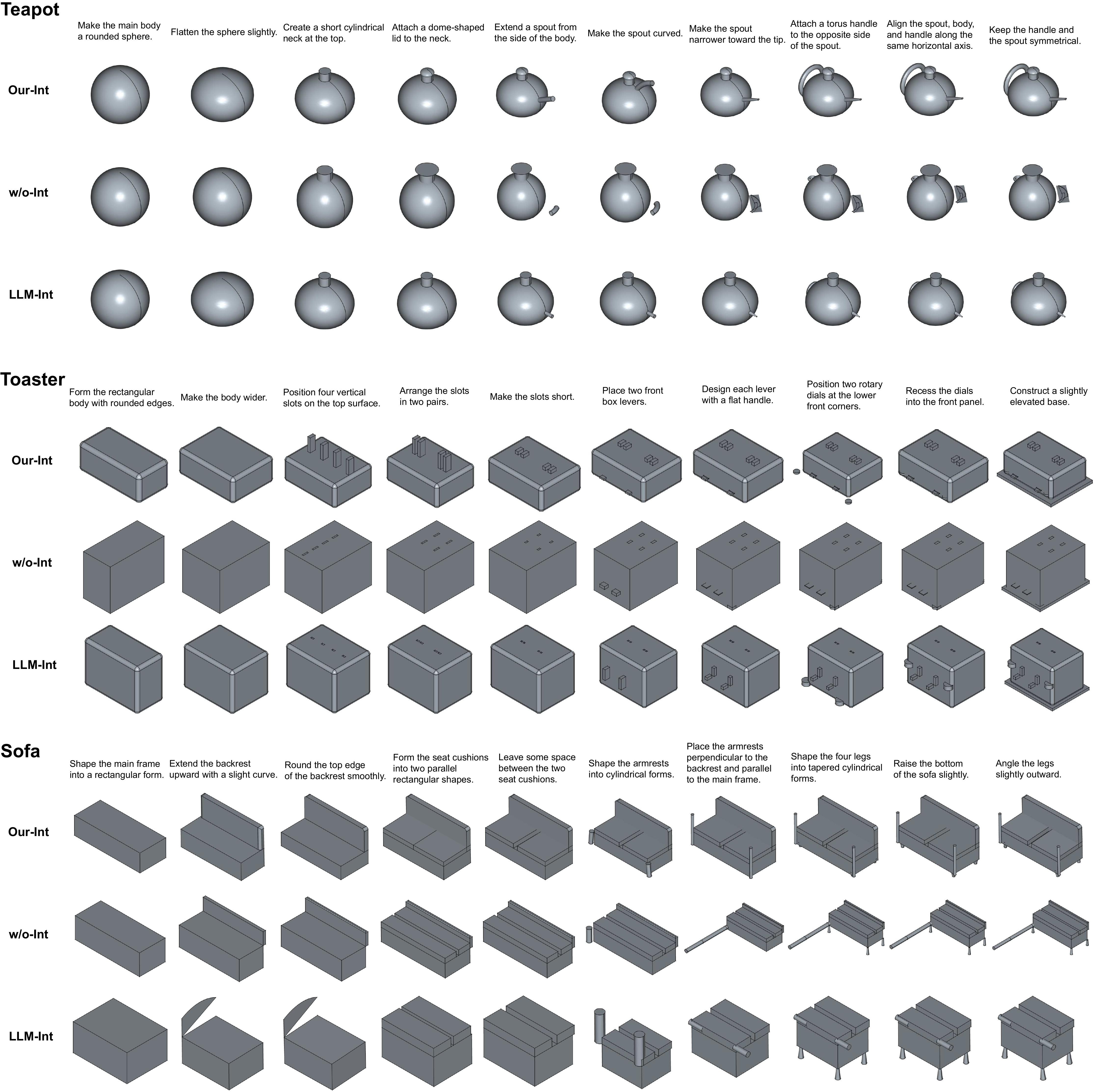}
    \vspace{-\baselineskip}
    \caption{\textbf{Step-by-step rendering results of some samples from the fast prototyping tasks}}
    \label{fig:supp-sbs}
\end{figure}

\subsection{Showcases of designers' instructions}\label{subsec:supp-exp-instruct}

There is a sample of ten-iteration instructions by one participant in the fast prototyping task of teapot design as follows.
\begin{lstlisting}[]
1. Make the main body a rounded sphere.
2. Flatten the sphere slightly.
3. Create a short cylindrical neck at the top.
4. Attach a dome-shaped lid to the neck.
5. Extend a spout from the side of the body.
6. Make the spout curved.
7. Make the spout narrower toward the tip.
8. Attach a torus handle to the opposite side of the spout.
9. Align the spout, body, and handle along the same horizontal axis.
10. Keep the handle and the spout symmetrical.
\end{lstlisting}

There is a sample of ten-iteration instructions by one participant in the fast prototyping task of toaster design as follows.
\begin{lstlisting}[]
1. Form the rectangular body with rounded edges.
2. Make the body wider.
3. Position four vertical slots on the top surface.
4. Arrange the slots in two pairs.
5. Make the slots short.
6. Place two front box levers.
7. Design each lever with a flat handle.
8. Position two rotary dials at the lower front corners.
9. Recess the dials into the front panel.
10. Construct a slightly elevated base.
\end{lstlisting}

There is a sample of ten-iteration instructions by one participant in the fast prototyping task of sofa design as follows.
\begin{lstlisting}[]
1. Shape the main frame into a rectangular form.
2. Extend the backrest upward with a slight curve.
3. Round the top edge of the backrest smoothly.
4. Form the seat cushions into two parallel rectangular shapes.
5. Leave some space between the two seat cushions.
6. Shape the armrests into cylindrical forms.
7. Place the armrests perpendicular to the backrest and parallel to the main frame.
8. Shape the four legs into tapered cylindrical forms.
9. Raise the bottom of the sofa slightly.
10. Angle the legs slightly outward.
\end{lstlisting}

\subsection{Showcases of interface DSL programs}\label{subsec:supp-exp-dsl}

There is a sample \ac{dsl} program collected from the fast prototyping of teapot design as follows.
\begin{lstlisting}[]
{
  "Parts": {
    "body": {
      "sphere_0": ["radius"]
    },
    "neck": {
      "cylinder_0": ["height", "radius", "diameter"]
    },
    "lid": {
      "sphere_0": ["height"]
    },
    "knob": {
      "sphere_0": ["height"]
    },
    "spout": {
      "cone_0": ["height", "angle", "radius"],
      "cylinder_1": ["diameter", "radius", "length", "height"]
    },
    "handle": {
      "torus_0": ["length", "height", "radius"]
    }
  },
  "Relationships": {
    "body <-> neck": ["top center"],
    "neck <-> lid": ["flush"],
    "lid <-> knob": ["top center"],
    "body <-> spout": ["side", "aligned_horizontal", "tilted_upward"],
    "body <-> handle": ["side", "opposite_spout", "aligned_horizontal", "higher"]
  }
}
\end{lstlisting}

There is a sample \ac{dsl} program collected from the fast prototyping of toaster design as follows.
\begin{lstlisting}[]
{
  "Parts": {
    "body": {
      "rounded_box_0": ["height", "length"]
    },
    "slot_1": {
      "rectangle_0": ["width", "length", "height"]
    },
    "slot_2": {
      "rectangle_0": ["width", "length", "height"]
    },
    "slot_3": {
      "rectangle_0": ["width", "length", "height"]
    },
    "slot_4": {
      "rectangle_0": ["width", "length", "height"]
    },
    "border_slot_1": {
      "thin_box_0": ["width", "length", "height"]
    },
    "border_slot_2": {
      "thin_box_0": ["width", "length", "height"]
    },
    "border_slot_3": {
      "thin_box_0": ["width", "length", "height"]
    },
    "border_slot_4": {
      "thin_box_0": ["width", "length", "height"]
    },
    "lever_1": {
      "curved_flat_handle_0": ["height", "radius"]
    },
    "lever_2": {
      "curved_flat_handle_0": ["height", "radius"]
    },
    "guide_rails_lever_1": {
      "vertical_thin_box_0": ["length", "height", "width"]
    },
    "guide_rails_lever_2": {
      "vertical_thin_box_0": ["length", "height", "width"]
    },
    "dial_1": {
      "cylinder_0": ["height", "radius"]
    },
    "dial_2": {
      "cylinder_0": ["height", "radius"]
    },
    "base": {
      "slightly_elevated_box_0": ["length", "width", "height"]
    },
    "crumb_tray": {
      "thin_box_0": ["length", "height", "width"]
    },
    "storage_area": {
      "cuboid_0": ["height", "radius"]
    },
    "power_cord": {
      "cylinder_0": ["height", "radius"]
    }
  },
  "Relationships": {
    "body <-> slot_1": ["top", "aligned"],
    "body <-> slot_2": ["top", "aligned"],
    "body <-> slot_3": ["top", "aligned"],
    "body <-> slot_4": ["top", "aligned"],
    "slot_1 <-> border_slot_1": ["surround"],
    "slot_2 <-> border_slot_2": ["surround"],
    "slot_3 <-> border_slot_3": ["surround"],
    "slot_4 <-> border_slot_4": ["surround"],
    "lever_1 <-> guide_rails_lever_1": ["beside"],
    "lever_2 <-> guide_rails_lever_2": ["beside"],
    "lever_1 <-> slot_1": ["aligned"],
    "lever_2 <-> slot_3": ["aligned"],
    "body <-> dial_1": ["front_bottom_corner", "recessed"],
    "body <-> dial_2": ["front_bottom_corner", "recessed"],
    "dial_1 <-> increments_dial_1": ["around"],
    "dial_2 <-> increments_dial_2": ["around"],
    "body <-> base": ["beneath", "contrast"],
    "base <-> crumb_tray": ["top", "slide_outward"],
    "body <-> storage_area": ["rear"],
    "storage_area <-> power_cord": ["extend_from"],
    "body <-> top_sides_base": ["seamless"]
  }
}
\end{lstlisting}

There is a sample \ac{dsl} program collected from the fast prototyping of wheelbarrow design as follows.
\begin{lstlisting}[]
{
  "Parts": {
    "basin": {
      "box_0": ["height", "radius"],
      "sphere_1": ["height", "radius"]
    },
    "front_edge": {
      "box_0": ["height", "radius"],
      "cylinder_1": ["height", "radius"]
    },
    "handles": {
      "cylinder_0": ["radius", "height"],
      "cylinder_1": ["radius", "height"]
    },
    "supports": {
      "cylinder_0": ["radius", "height"],
      "cylinder_1": ["radius", "height"]
    },
    "frame": {
      "u_shape_frame_0": ["radius", "height"]
    },
    "wheels": {
      "cylinder_0": ["radius", "height"],
      "cylinder_1": ["radius", "height"]
    },
    "axle": {
      "cylinder_0": ["radius", "height"]
    }
  },
  "Relationships": {
    "basin <-> front_edge": ["sloped_downward"],
    "basin <-> handles": ["attached_to_rear"],
    "handles <-> supports": ["connected"],
    "frame <-> basin": ["underneath"],
    "frame <-> wheels": ["symmetrical"],
    "wheels <-> axle": ["through"],
    "axle <-> frame": ["horizontal"],
    "axle <-> basin": ["behind_front_edge"],
    "handles <-> frame": ["rear_higher"]
  }
}
\end{lstlisting}

\subsection{Success and failure cases}\label{sec:supp-exp-case}

Success cases demonstrate the advancements of our interface and failure cases shape its boundary. 

\begin{figure}[ht!]
    \centering
    \includegraphics[width=\linewidth]{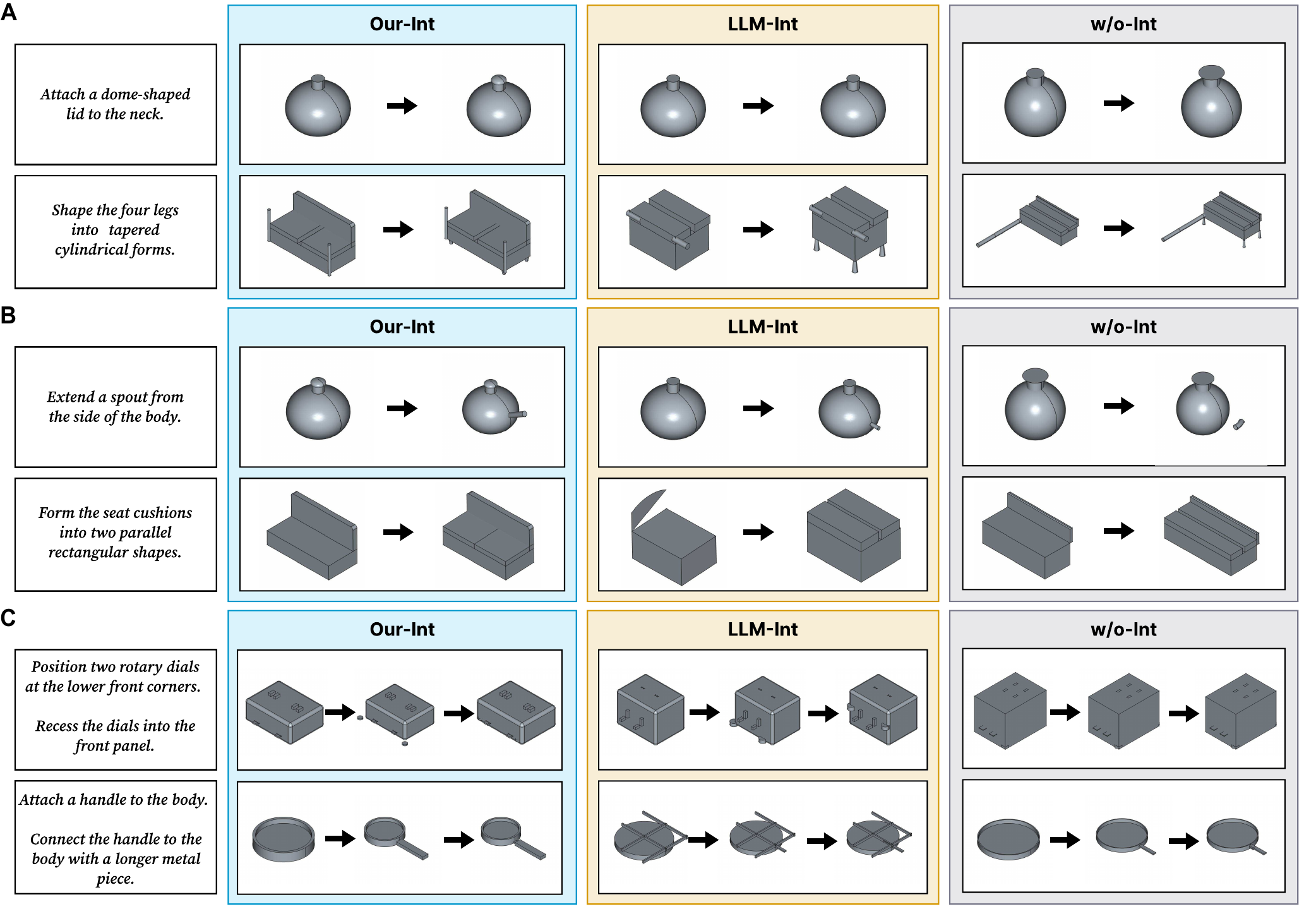}
    \vspace{-\baselineskip}
    \caption{\textbf{Demonstrations of success cases}}
    \label{fig:supp-success-cases}
\end{figure}

Success cases: (i) \texttt{Our-Int} effectively maintains and guides the transformation of basic shapes and their modifications, mapping abstract designer instructions to concrete shape changes (see \cref{fig:supp-success-cases}A); (ii) \texttt{Our-Int} can automatically infer a commonsense spatial distribution of components when designer instructions lack explicit spatial constraints (see \cref{fig:supp-success-cases}B); and (iii) \texttt{Our-Int} translates modifications at a finer granularity, identifying which component and which attributes are affected (see \cref{fig:supp-success-cases}C). 

\begin{figure}[ht!]
    \centering
    \includegraphics[width=\linewidth]{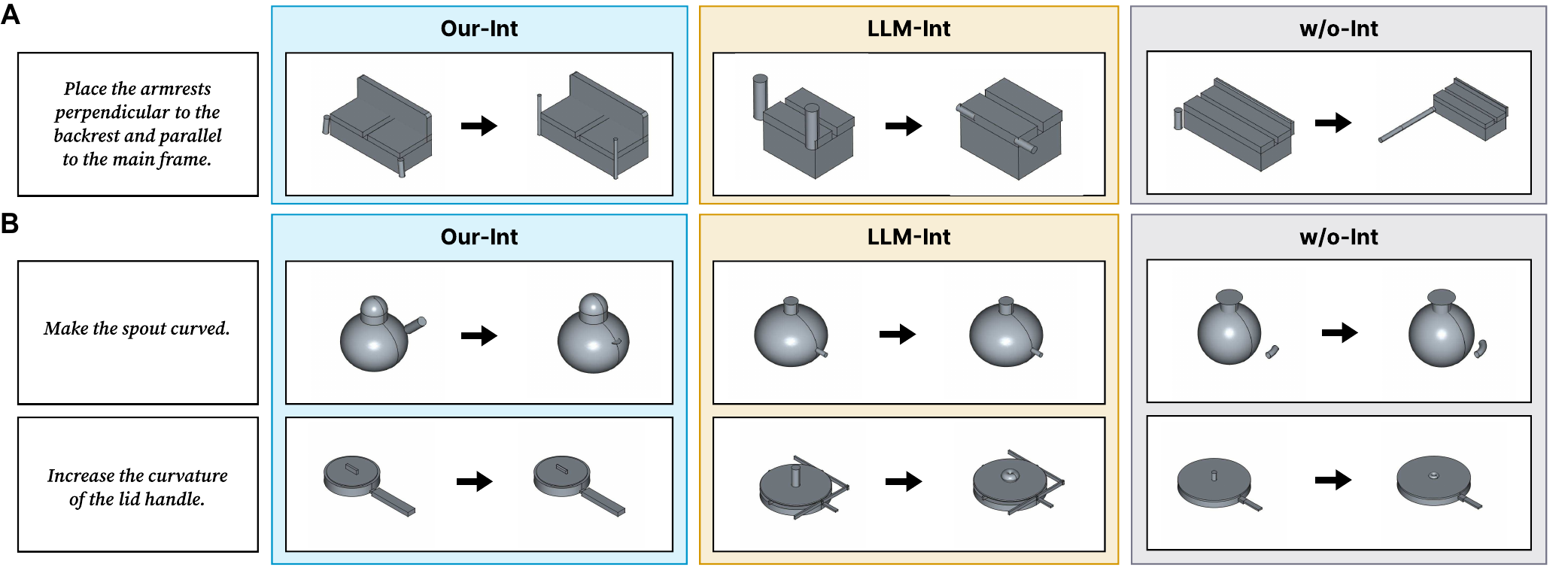}
    \vspace{-\baselineskip}
    \caption{\textbf{Demonstrations of failure cases}}
    \label{fig:supp-failure-cases}
\end{figure}

Failure cases: (i) for qualitative spatial constraints like ``vertical'' or ``parallel'', \acp{llm} sometimes fail to map them correctly to precise positions and orientations due to their weak spatial reasoning (see \cref{fig:supp-failure-cases}A); and (ii) for certain abstract and complex instructions---such as operations involving Bezier curves---current methods sometimes fail to capture the correct approach (see \cref{fig:supp-failure-cases}B).

\section{Implementation details}\label{sec:supp-implement}

\subsection{Computational cost}\label{sec:supp-implement-cost}

We leverage OpenAI's GPT-4o API as the backbone \ac{llm} for both domain adaptation and runtime execution of the interface. Automatically designing a domain-specific interface incurs an average cost of approximately \$10 per domain, with full adaptation across eight distinct domains. Operational costs during prototyping remain economical: executing the interface for targeted control consumes \$0.3 per ten refinement iterations. 

\subsection{Implementation of MCMC sampling}\label{sec:supp-implement-mcmc}

The \ac{mcmc} sampling process for construct expansion leverages \acp{llm} as a dynamic proposal generator. For each seed construct $L^{(t)}$, the proposal distribution $q(L' \mid L^{(t)})$ is instantiated through \ac{llm} prompts that request domain-aware perturbations, such as: \emph{``Modify [current construct: \texttt{handle length}] to explore ergonomic variations, considering material constraints.''} The \ac{llm} generates candidate perturbations $L'$, which are parsed into valid \ac{dsl} construct samples. The acceptance probability is estimated using the \ac{llm}'s likelihood scores for $L'$ versus $L^{(t)}$ under domain-specific contexts. To ensure detailed balance, we maintain a buffer of rejected proposals for delayed acceptance. By initializing $M$ parallel chains and performing $N$ steps per chain, the sampling process efficiently explores the combinatorial space of $\mathcal{K}$ while preserving semantic diversity across domains.

\subsection{Feasibility validation via documentation-guided RAG}\label{sec:supp-implement-rag}

The validation phase employs a \ac{rag} system built from the modeling engine's programmatic documentation. First, we generate structured documentation by parsing the FreeCAD library's source code (hosted at \url{https://wiki.freecad.org/Category:Developer_Documentation}) using Doxygen (as suggested by the FreeCAD official website), extracting function signatures, parameter constraints, and geometric operation specifications. This documentation is chunked into text segments and indexed in a vector store. During validation, each interface construct is converted to a natural language query (\eg, \emph{``Adjust cylindrical surface diameter''}) and used to retrieve the top-$k$ relevant documentation entries. This \ac{rag}-driven process ensures the interface remains grounded in the modeling engine's actual capabilities, closing the loop between conceptual abstraction and technical executability.  

Specifically, the mechanism of feedback generation evaluates constructs through both \ac{llm}-as-a-judge analysis and \ac{cad} engine constraints~\citep{zheng2023judging}. It assesses three feasibility aspects and provides feedback to refine heuristics for subsequent iterations, all with minimal human intervention. First, designer language constructs (\eg, a ``ring-shaped teapot body'') cannot be translated into modeling operations. If a construct is unsupported, heuristics such as pruning incompatible geometric primitives are applied and the LLM is prompted to propose alternative base shapes (\eg, torus segments, since CAD engines do not directly support a ``ring''). A high frequency of such cases indicates that the constructs are overly abstract. Second, modeling constructs (\eg, sofa cushions formed by fusing two cylinders) lack equivalent high-level design terms. For missing constructs, heuristics like generating composite-shape directives (\eg, ``create cushion from combined cylinders'') are triggered in the next sampling iteration. A high occurrence of these cases suggests insufficient diversity in the designer’s language. Third, no overlap between the finest designer constructs and the coarsest modeling operations. For mismatches (\eg, unsupported ``material texture'' operations), heuristics such as incompatibility pruning are employed, permanently removing non-viable constructs.

\subsection{Updating of iteration}\label{sec:supp-implement-updating}

The optimization of \cref{eq:m-step} is achieved through iterations alternating construct expansion and feasibility validation. During the construct expansion phase, the heuristics adjust exploration strategies based on the interface’s state and feedback from prior iterations. When diversity is insufficient (\eg, limited variation in designs), the heuristics broaden exploration breadth by prompting the \ac{llm} with directives like \emph{``generate diverse handle configurations for teapots''}. Conversely, if diversity is high, the heuristics narrow exploration breadth by prompting with constraints. When constructs are overly abstract (\eg, \emph{``refine shape''}), heuristics increase exploration depth by decomposing the constructs into atomic operations. If constructs are excessively granular, the heuristics reduce exploration depth by encapsulating low-level commands into functions (\eg, \emph{``smooth contour''}). In the feasibility validation phase, heuristics enforce alignment with the modeling engine’s capabilities by pruning constructs incompatible with \ac{cad} engine's constraints. This is now discussed in the revised manuscript.

Three intermediate metrics are leveraged to monitor the iterative process: (i) soundness, ensuring all language constructs used by designers can be implemented in the modeling process; (ii) completeness, ensuring all modeling process constructs are represented in designers' language; (iii) granularity alignment, ensuring proper overlap between the finest-grained constructs in designers' language and the coarsest-grained constructs in the modeling process. 

\begin{figure}[ht]
    \centering
    \includegraphics[width=\linewidth]{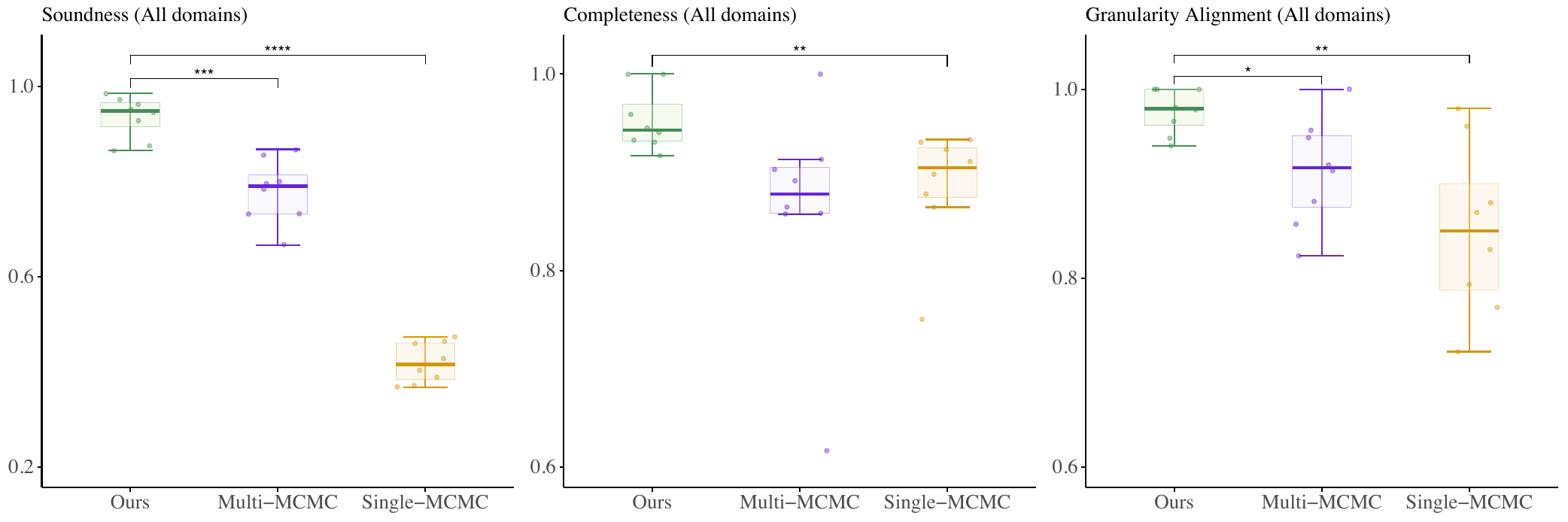}
    \vspace{-\baselineskip}
    \caption{\textbf{Quantitative results of the intermediate metrics}}
    \label{fig:supp-inter-res}
\end{figure}

Under these intermediate metrics, we explore several alternative interfaces during the iterative process. The interfaces we examined are framed as (i) Single-MCMC: single-scale sampling without updating, sampling and clustering constructs within a single chain but lacks multi-chain diversity; (ii) Multi-MCMC: multi-scale sampling with updating, exploring diverse constructs via parallel chains but omits iterative optimization; and (iii) Ours: extending the previous two alternative interfaces by integrating multi-scale sampling and converged iterative refinement. Quantitative results support the design rationale of our interface (see \cref{fig:supp-inter-res}). This superior systematic representation directly contributes to the improved final performance when integrated with an \ac{llm} for \ac{cad}. 

\section{Reproducibility}\label{sec:supp-reproduce}

The project page with supplementary files for reproducing the results of this paper will be available at \url{https://autodsl.org/concept/papers/icml25shi.html}.

\section{Limitations}\label{sec:supp-limit}

As a representation of interface designed for a relatively under-researched problem, the design and evaluation of the proposed methodology come with limitations, leading to further investigations: 
\begin{itemize}[noitemsep,nolistsep,topsep=0pt,leftmargin=*]
    \item We currently model the interaction between industrial designers and the interface as a Markov Process, where each step of adjustment is only conditioned on the last step. Can we model the interactive process in a more sophisticated fashion, thereby taking the temporal dynamics along the process into consideration, such as Bayesian updating, working memory, and the Aha! moment?
    \item We presently regard \ac{nl} instructions as the vehicle conveying the intentions of industrial designers. Can we include more types of medium for industrial designers' ideas, such as primal sketches on 2D plane, gesture trajectories in 3D space, or digital clays with surfaces covered by tactile sensors in a higher degree-of-freedom, as the input of our proposed interface?
    \item Creating groundtruth-level modeling commands for target products requires the dedication of considerable efforts of experts. Can we employ an approach with a higher extent of automation to reduce the cost groundtruth annotation, thereby broadening the scope of the products in the testing set, toward some small group domains of product?
\end{itemize}
With many questions unanswered, we hope to explore more on building interface for the targeted control of fast prototyping and beyond.

\end{document}